\documentclass[10pt,twocolumn,letterpaper]{article}

\usepackage{multirow}
\usepackage{lineno}
\usepackage{amsmath}       %
\usepackage{amssymb}       %
\usepackage{bbding}        %
\usepackage[dvipsnames,table,xcdraw]{xcolor} 
\definecolor{mygray}{gray}{0.9} %
\newcommand{\red}[1]{{\color{red}{#1}}}

\usepackage{xspace}
\newcommand{\app}{\raise.17ex\hbox{$\scriptstyle\sim$}}
\makeatletter\renewcommand\paragraph{\@startsection{paragraph}{4}{\z@}
  {.15em \@plus1ex \@minus.2ex}{-.5em}{\normalfont\normalsize\bfseries}}\makeatother
\usepackage[pagenumbers]{cvpr} %

\definecolor{cvprblue}{rgb}{0.21,0.49,0.74}
\usepackage[pagebackref,breaklinks,colorlinks,allcolors=cvprblue]{hyperref}

\def\paperID{13893} %
\def\confName{CVPR}
\def\confYear{2025}

\title{CLIP\texttt{S}: An Enhanced CLIP Framework for Learning with Synthetic Captions}

\author{Yanqing Liu$^1$ \quad Xianhang Li$^1$ \quad Zeyu Wang$^1$ \quad Bingchen Zhao$^2$ \quad Cihang Xie$^1$\vspace{.3em}\\
$^1$ UC Santa Cruz \quad $^2$ University of Edinburgh \vspace{.3em} \\
Project Page: \url{https://ucsc-vlaa.github.io/CLIPS/}
}

\begin{document}
\maketitle
\begin{abstract}

Previous works show that noisy, web-crawled image-text pairs may limit vision-language pretraining like CLIP and propose 
learning with synthetic captions as a promising alternative. 
Our work continues this effort, introducing two simple yet effective designs to better leverage richly described synthetic captions. Firstly, by observing a strong inverse effect in learning with synthetic captions---the short synthetic captions can generally lead to MUCH higher performance than full-length ones---we therefore fed only partial synthetic captions to the text encoder.  Secondly, we incorporate an autoregressive captioner to mimic the recaptioning process---by conditioning on the paired image input and web-crawled text description, the captioner learns to predict the full-length synthetic caption generated by advanced MLLMs.
Experiments show that our framework significantly improves zero-shot performance in cross-modal retrieval tasks, setting new SOTA results on MSCOCO and Flickr30K. Moreover, such trained vision encoders can enhance the visual capability of LLaVA, showing strong improvements on a range of MLLM benchmarks. 
\end{abstract}
    
\section{Introduction}
\label{sec:intro}

The availability of large-scale image-text datasets, such as LAION~\cite{schuhmann2022laion} and DataComp~\cite{gadre2024datacomp}, has been a key driver of the rapid development of vision-language models in recent years~\cite{radford2021learning, yu2022coca, li2022blip, bai2023qwen, chen2024internvl, liu2024improved}.  Nonetheless, these web-crawled datasets are generally noisy and not-high-quality (\eg, image-text pairs could be mismatched), which potentially limits further performance improvements~\cite{jia2021scaling}. Consequently, many works seek to improve the dataset quality by re-generating paired textual descriptions using multimodal large language models (MLLM) and incorporating these synthetic captions into training \cite{fan2024improving, nguyen2024improving, lai2025veclip}.

A straightforward approach to learning with synthetic captions is to simply replace the raw, web-crawled captions with rewritten ones~\cite{nguyen2024improving, li2024if, lai2025veclip}. As demonstrated in VeCLIP~\cite{lai2025veclip} and Recap-DataComp-1B~\cite{li2024if}, (partially) substituting the original captions with those generated by advanced MLLMs during CLIP training can substantially enhance the models' capabilities, especially in cross-modal retrieval tasks. Building upon this line of research, our work continues the exploration of training with synthetic captions but focuses on enhancing the vision-language pretraining framework to better leverage these captions, similar to prior efforts~\cite{fan2024improving, lai2025veclip, zheng2025dreamlip}.

\begin{figure}[t!]
    \centering
    \includegraphics[width=\linewidth]{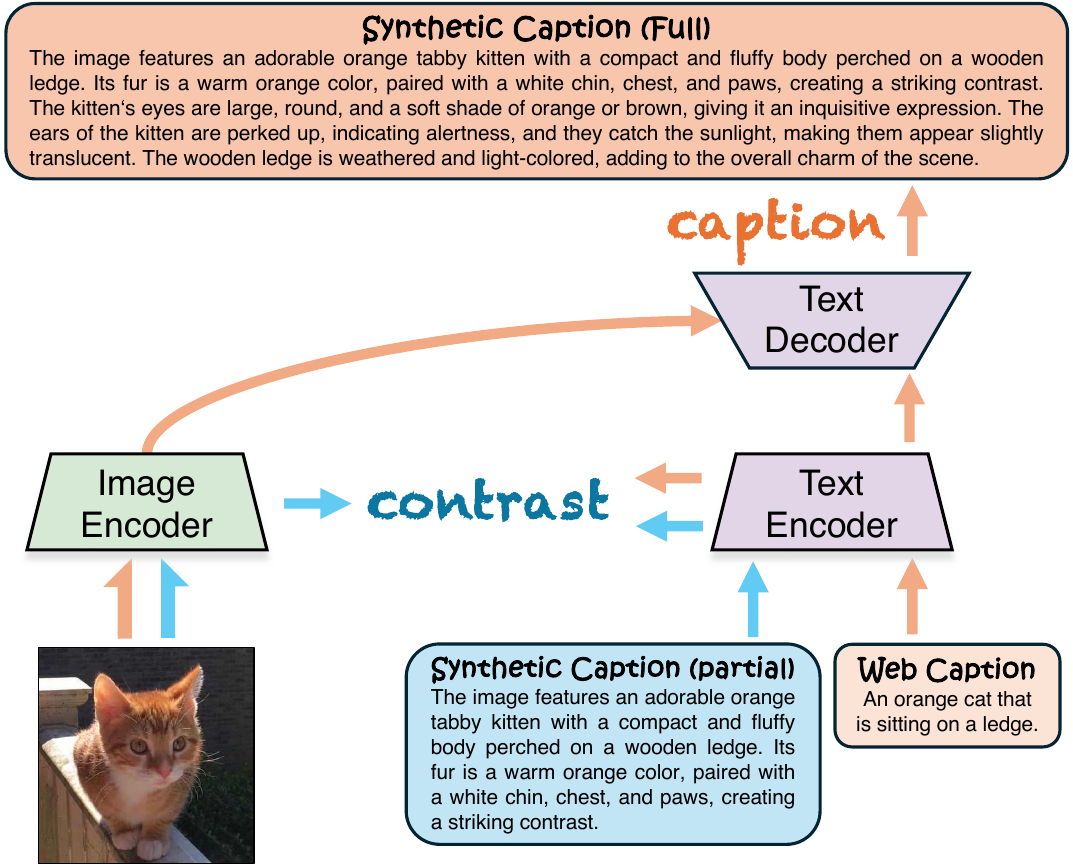}
    \caption{The pipeline of our proposed CLIP\texttt{S}. We introduce two simple yet effective designs---1) only a subpart of the synthetic caption is used in contrastive learning and 2) a captioner to predict the full synthetic caption based on the web-crawled caption and the image---to better leverage synthetic captions. Our method registers new SOTA results on MSCOCO, achieving 76.4\% in text retrieval and 57.2\% in image retrieval.}
    \label{fig:pipeline}
\end{figure}

Because synthetic captions are typically highly descriptive---much longer and containing more detailed information than web-crawled captions---we introduce two simple yet effective designs to better leverage them in CLIP training. Our first design, inspired by the inverse scaling law of CLIP training revealed in~\cite{li2024inverse}, involves randomly sampling a portion of the synthetic caption to serve as input to the text encoder. Interestingly, we observe that transitioning from web-crawled captions to synthetic captions leads to a tipping point in these inverse effects---rather than hurting performance (with larger models being less affected with shorter captions) as noted in~\cite{li2024inverse,li2023clipa}, dropping parts of the synthetic caption surprisingly leads to performance improvements. The strongest performance is achieved when we randomly select a single sentence from the synthetic caption as the paired text description for contrastive learning, discarding the rest. Moreover, as shown in \cite{li2024inverse}, learning with shorter text reduces overall computation, providing additional benefits.

Since the synthetic caption is only partially used in contrastive learning, our second design aims to incorporate their \textit{full} use in an auxiliary task. Specifically, we follow CoCa~\cite{yu2022coca} by incorporating an autoregressive decoder to predict captions. However, unlike the symmetric design in CoCa, where the input text and the output text are the same (\ie, the web-crawled caption), we introduce an asymmetric design: the input to the text decoder is the web-crawled caption, and the prediction target is the full-length synthetic caption. This learning behavior mimics the recaptioning process performed by MLLMs and ensures full utilization of the knowledge within the complete synthetic captions.

We termed this resulting training framework as CLIP\texttt{S}.  Experimental results show that CLIP\texttt{S} significantly enhances zero-shot performance in cross-modal retrieval. For example, with a ViT-L backbone, our CLIP\texttt{S} substantially outperforms SigLip \cite{zhai2023sigmoid} by 4.7 (from 70.8 to 75.5) on MSCOCO's R@1 text retrieval, and by 3.3 (from 52.3 to 55.6) on MSCOCO's R@1 image retrieval. With increased computational resources and scaling, our best model further achieves 76.4\% and 96.6\% R@1 text retrieval performance on MSCOCO and Flickr30K respectively, and 57.2\% and 83.9\% R@1 image retrieval performance on the same datasets, setting new state-of-the-art (SOTA) results. Moreover, our CLIP\texttt{S} framework contributes to building stronger MLLMs---replacing the visual encoder from OpenAI-CLIP \cite{radford2021learning} with our CLIP\texttt{S} in LLaVA \cite{liu2024improved,liu2024visual} leads to strong performance gains across a range of MLLM benchmarks.

\section{Related Works}
\label{sec:related}
\paragraph{Vision-language Pre-training}
Vision-language pre-training aims to align vision and language modalities to create a unified representation for various image-text understanding tasks. Existing frameworks predominantly adopt either single-stream architecture \cite{li2019visualbert,chen2020uniter,kim2021vilt}, which jointly represents different modalities using a shared encoder, or two-stream architectures \cite{lu2019vilbert,li2021align,radford2021learning,jia2021scaling}, which employ two independent encoders to process visual and textual inputs separately.
Our work builds upon the latter approach, specifically focusing on CLIP, which leverages contrastive loss to align image and text representations.

The further enhancements to CLIP have generally pursued two main directions. The first direction involves extending CLIP's capabilities into generative tasks,  
such as image captioning, visual question answering, and image grounding. 
Notable works in this vein include  CoCa~\cite{yu2022coca} and BLIP~\cite{li2022blip}, which  enables the unification of image-text understanding and generation tasks by transitioning from an encoder-only architecture to an encoder-decoder architecture. The second direction focuses on optimizing 
vision-language contrastive learning. FILIP~\cite{yao2021filip} mitigates the fine-grained alignment issue in CLIP by modifying the contrastive loss. SigLip~\cite{zhai2023sigmoid} replaces contrastive loss with sigmoid loss to optimize computation efficiency. Llip~\cite{lavoie2024modeling} models captions diversity by associating multiple captions with a single image representation. CLOC~\cite{chen2024contrastive} enhances regional localization by introducing a region-text contrastive loss, improving the model's ability to focus on specific image regions corresponding to textual inputs. Our work also aims to improve CLIP but focuses on enhancing the leverage of richly described synthetic captions in training.

\paragraph{Learning from Synthetic Captions}
Recognizing that web-crawled image-text datasets are often noisy and contain mismatched pairs, recent works seek to improve dataset quality by rewriting captions. ALip~\cite{yang2023alip} uses the OFA~\cite{wang2022ofa} model to generate synthetic captions and introduces a bi-path model to integrate supervision from two types of text. LaCLIP~\cite{fan2024improving} rewrites captions using LLMs such as ChatGPT~\cite{openai2022chatgpt} and randomly selects one of these rephrasings during training. Nguyen \etal \cite{nguyen2024improving} use BLIP-2~\cite{li2023blip} to rewrite captions for image-text pairs with low matching degrees in the original dataset. VeCLIP~\cite{lai2025veclip} first uses LLaVA~\cite{liu2024visual} to generate synthetic captions with rich visual details, then uses an LLM to fuse the raw and synthetic captions. Liu \etal \cite{liu2023mllms} propose using multiple MLLMs to rewrite captions and apply text shearing to improve caption diversity. ShareGPT4V~\cite{chen2023sharegpt4v} feeds carefully designed prompts and images to GPT-4V~\cite{achiam2023gpt} to create a high-quality dataset, which has been widely adopted in subsequent works~\cite{lin2024moe, chen2024pixart, chu2024mobilevlm}. Li \etal \cite{li2024if} rewrites captions using a more advanced LLaMA-3-based~\cite{meta2024introducing} LLaVA in the much larger-scale DataComp-1B dataset~\cite{gadre2024datacomp}. SynthCLIP~\cite{hammoud2024synthclip} trains entirely on synthetic datasets by generating image-text pairs using text-to-image models and LLMs. DreamLip~\cite{zheng2025dreamlip} builds a short caption set for each image and computes multi-positive contrastive loss and sub-caption specific grouping loss. Our first design is closely related to DreamLip, but presents a more general message about the inverse effect of learning with synthetic captions (\ie, short ones are highly preferred). Additionally, to fully leverage the information in the complete synthetic captions, our second design incorporates an CoCa-like but asymmetric decoder (\ie, web-crawled caption as the input, full synthetic caption as the output).

\begin{figure*}[h!]
    \centering
    \includegraphics[width=\textwidth]{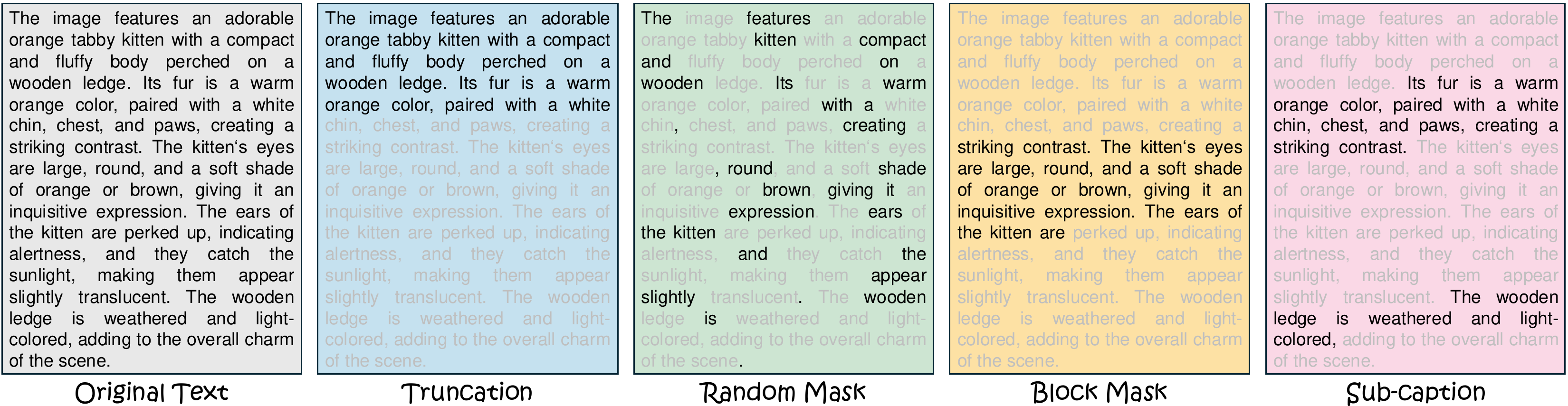}
    \vspace{-1.3em}
    \caption{Visualization of four different token reduction strategies. These strategies can improve the model's learning efficiency on synthetic captions to varying degrees. Among these strategies, the sub-caption and block mask perform best.}
    \label{fig:mask_strategy}
    \vspace{-.5em}
\end{figure*}
\section{Method}
\label{sec:method}
This section first covers related preliminaries, including the contrastive loss in CLIP and the generative loss in CoCa. We then present the observed inverse effect of learning with synthetic captions. Lastly, we introduce our simplified multi-positive contrastive loss for the encoder and an asymmetric decoder design for predicting full-length synthetic captions.

\subsection{Preliminaries}
\paragraph{CLIP} Let \(\{(x_i, y_i)\}_{i=1}^N\) denote a set of image-text pairs, where \(x_i\) represents an image and \(y_i\) represents the corresponding text description. Images and texts are encoded using a vision encoder \(f(\cdot)\) and a text encoder \(g(\cdot)\), respectively. 
For the image loss \(L_{\text{I}}\), we first calculate the similarity of each image in the local batch with all texts. These similarity values are then used to compute the InfoNCE loss~\cite{oord2018representation}, with correctly matched image-text pairs treated as positive samples and non-matching pairs as negative samples. The formula can be expressed as:
\begin{equation}
\label{equation:contrstive_loss}
L_{\text{I}} = - \frac{1}{N} \sum_{i=1}^{N} \log \frac{\exp\left(\text{sim}(f(x_i), g(y_i)) / \tau\right)}{\sum_{j=1}^{N} \exp\left(\text{sim}(f(x_i), g(y_j)) / \tau\right)}
\end{equation}
where \(\text{sim}(f(x_i), g(y_j))\) denotes the similarity function (e.g., cosine similarity) and \(\tau\) is a temperature parameter.

For the text loss \(L_{\text{text}}\), the formula is similar, but the roles of images and texts are swapped. The total loss \(L_{\text{contrast}}\) is the average of the image loss and text loss:
\begin{equation}
    L_{\text{contrast}} = \frac{1}{2} \left( L_{\text{image}} + L_{\text{text}} \right)
\end{equation}

\paragraph{CoCa} The generative loss here is instantiated as autoregressively predicting the next token in a target sequence conditioned on image features and previously generated tokens. Specifically, the text branch follows an encoder-decoder architecture where the unimodal text encoder encodes text, and the multimodal text decoder generates the output sequence. This process can be formalized as:
\begin{equation} 
\label{equation:generative loss}
L_{\text{gen}} = - \sum_{t=1}^{T} \log P(y_t \mid y_{<t}, E(x))
\end{equation}
where \(y_t\) represents the target token at time step \(t\), \(y_{<t}\) denotes all tokens preceding \(y_t\), \(E(x)\) represents the encoded features of the input \(x\), and \(P(y_t \mid y_{<t}, E(x))\) is the conditional probability of the target token \(y_t\) given the previous tokens and encoded features.
\begin{figure*}[h!]
    \centering
    \begin{subfigure}[b]{0.24\textwidth}
        \centering
        \includegraphics[width=\textwidth]{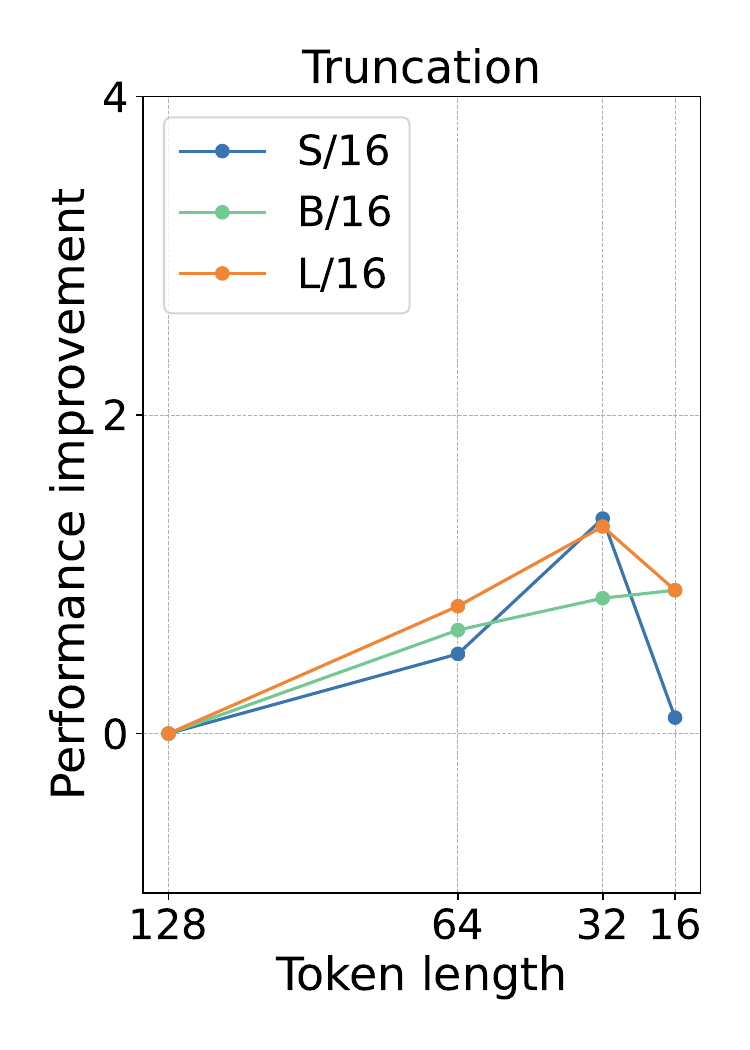}
    \end{subfigure}
    \hfill
    \begin{subfigure}[b]{0.24\textwidth}
        \centering
        \includegraphics[width=\textwidth]{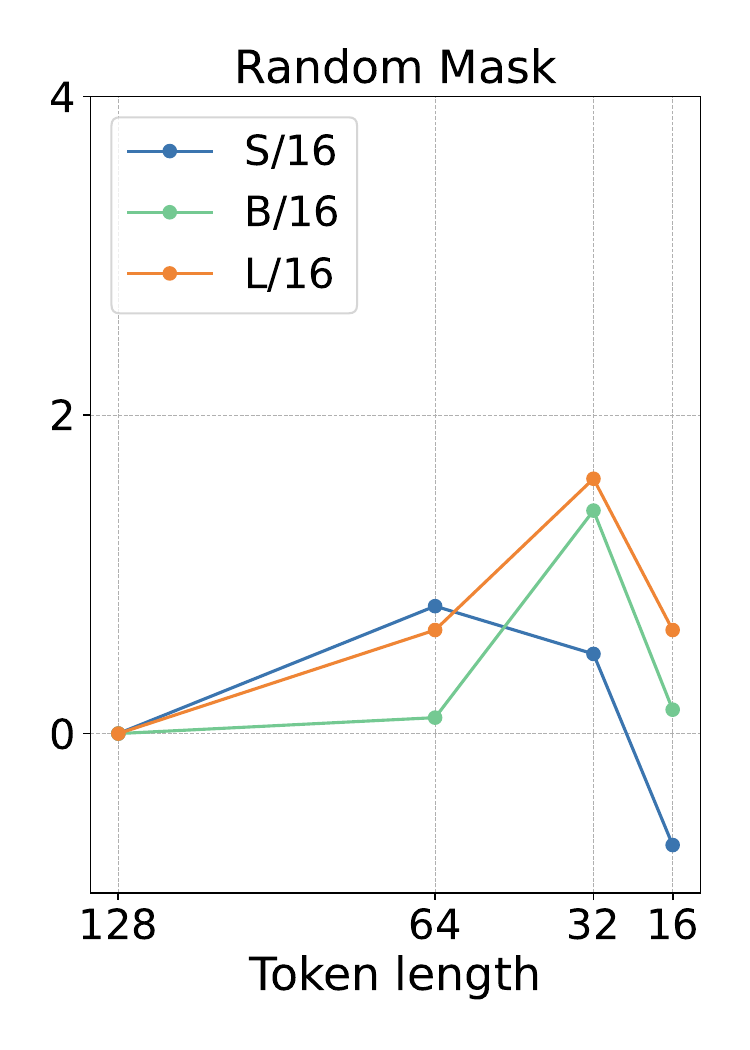}
    \end{subfigure}
    \hfill
    \begin{subfigure}[b]{0.24\textwidth}
        \centering
        \includegraphics[width=\textwidth]{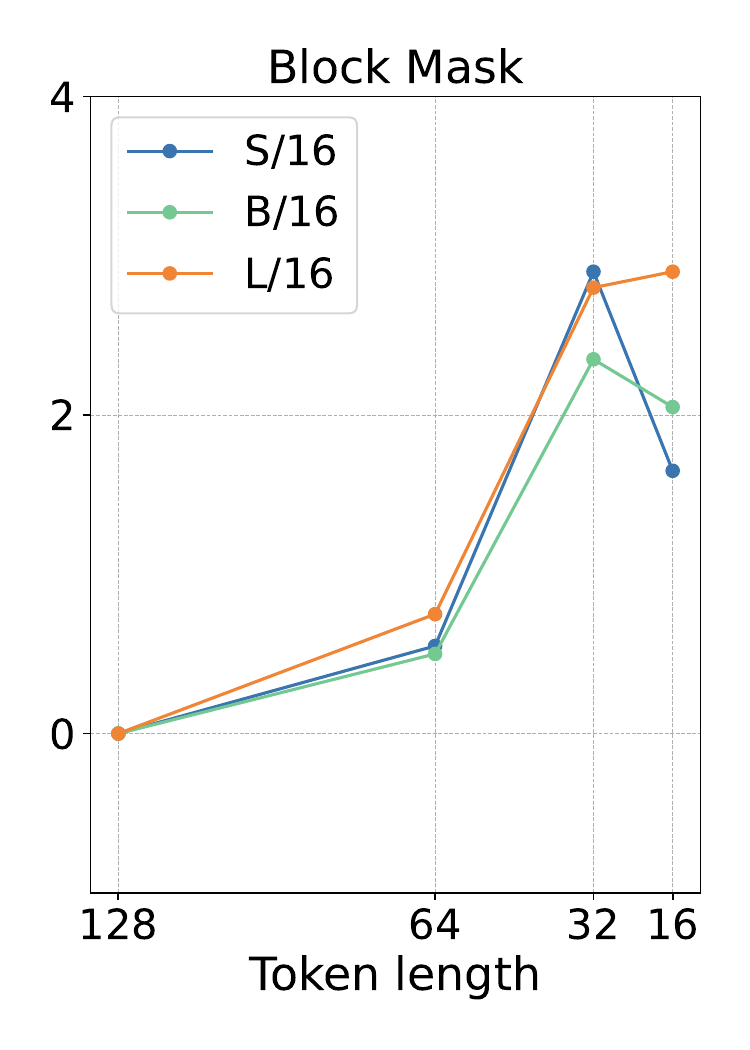}
    \end{subfigure}
    \hfill
    \begin{subfigure}[b]{0.24\textwidth}
        \centering
        \includegraphics[width=\textwidth]{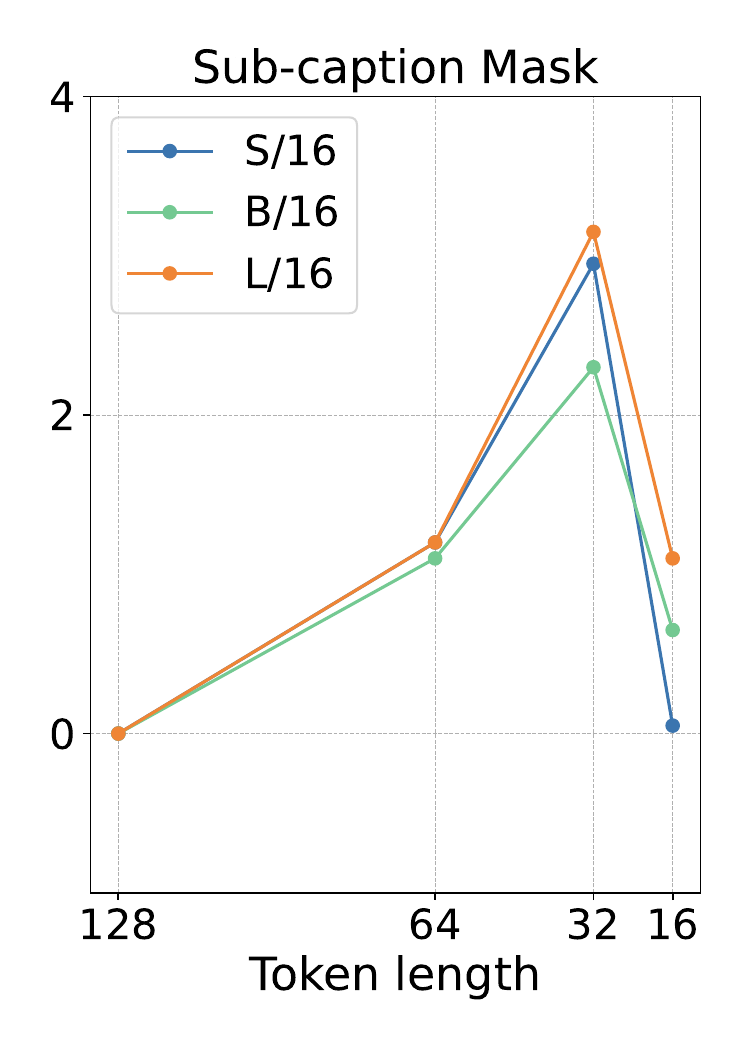}
    \end{subfigure}
    \vspace{-1.3em}
    \caption{The inverse scaling effect of synthetic captions. Unlike the performance drop from reducing token length in original captions, shortening the token length of synthetic captions consistently improves model performance.}
    \vspace{-.5em}
    \label{fig:scaling law}
\end{figure*}

\subsection{Inverse Effect with Synthetic Captions}
\label{sec:inverse_scaling_law}
Inspired by the prior work \cite{li2024inverse}, we first check how CLIP models behave when learning with \textit{shorter synthetic captions}.
As illustrated in Figure~\ref{fig:mask_strategy}, we employ four strategies to reduce the token length of synthetic captions: truncation, random mask, block mask, and sub-caption mask. The truncation, random mask, and block mask strategies are based on the approaches described in~\cite{li2024inverse}. Notably, we omit the syntax mask from~\cite{li2024inverse}, which performed best when learning with web-crawled captions but yielded very non-competitive results in our synthetic caption experiments.\footnote{We tested the syntax mask with the ViT-B/16 model and a token length of 32, finding that it achieved an average retrieval performance of 52.8, slightly below the performance without any mask. In contrast, under the same setup, the truncation, random mask, and block mask strategies led to noticeable performance improvements, as shown in Figure~\ref{fig:scaling law}.} The sub-caption mask strategy is adapted from the sub-caption extraction method proposed in~\cite{zheng2025dreamlip}.

Specifically, given a synthetic caption sequence of length \(K\) and a target token length \(L\), the truncation directly selects the first \(L\) tokens; The random mask obtains \(L\) tokens through random sampling; The block mask randomly selects a starting point in the original sequence and take the subsequent \(L\) tokens. For the sub-caption mask, we split the synthetic caption at periods, resulting in segments \{\( S_1, S_2, \dots, S_n \)\}, where \(n\) is the number of sub-captions. 
We randomly select a sub-caption and check its length: If the length meets and exceeds the predefined limit, we truncate it; Otherwise, we randomly select another sub-caption from the remaining ones, concatenate it with the previous segment, and then check the length again. This process can be formalized as:
\begin{equation}
    \text{Subcaption}(S, L) = 
\begin{cases} 
    \text{Truncate}(S_i), & \text{if } |S_i| > L \\
    \text{Concat}(\{S_i, S_j, \dots\}), & \text{if } |S_i| < L 
\end{cases}
\end{equation}
where \( S_i \) and \( S_j \) denotes randomly selected sub-captions from the set \(\{S_1, S_2, \dots, S_n\}\). 

We follow the training setup described in~\cite{li2024inverse} to train all models. Additionally, similar to~\cite{zheng2025dreamlip}, we include both synthetic captions and the original web-crawled captions in the contrastive learning process. We conduct experiments using ViT~\cite{dosovitskiy2020image} models of sizes S/16, B/16, and L/16 for each token reduction strategy, progressively reducing the token length from 128 to 64, 32, and 16. We evaluate the models' average R@1 retrieval performance on the MSCOCO dataset and present the results in Figure~\ref{fig:scaling law}.

\paragraph{Main observations.} Firstly, these results confirm that the inverse effects also exists when training with synthetic captions; that is, learning with reduced token lengths is generally preferred. But we stress that this inverse effect becomes significantly more pronounced when transitioning from web-crawled captions to synthetic captions---while the inverse effect in \cite{li2024inverse} refers to that larger models being less \textit{adversely} affected when learning with reduced-length web-crawled captions, our experiments hereby reveals that, when using synthetic captions, reducing the token length consistently yields \textit{noticeable performance gains} across all model sizes.
In other words, learning with synthetic captions at a reduced length is a more ``optimal'' way to train CLIP, enjoying the benefits of both higher performance and higher efficiency (due to less text token used). 
As a side note, this observation also corroborates the prior works on showing text shearing~\cite{liu2023mllms} and sub-caption extraction~\cite{zheng2025dreamlip} are effective strategies to process long synthetic captions for enhancing performance.

Moreover, by taking a closer look at Figure \ref{fig:scaling law}, we note that
the sub-caption mask and block mask perform the best, and the truncation and random mask are slightly less effective. We conjecture that, in addition to sequence length being a critical factor in contrastive learning, the diversity and coherence of the reduced text segments also play significant roles. Furthermore, although we do not observe a clear scaling trend between the S/16 and B/16 models, it is evident that the larger L/16 model achieves the most substantial performance gains compared to the smaller models.

\subsection{Encoder: Learning with Short Captions}
\label{sec:encoder}
Based on the observation above, we adopt the sub-caption strategy as the default preprocessing method for synthetic captions in our encoder design. Specifically, considering that all ViTs generally achieve the strongest performance at an input token length of 32, roughly corresponding to one or two sentences from the full synthetic caption, we thereby implement the sub-caption strategy by sampling a single random sentence from each full synthetic caption.  This sampled sentence, along with the original web-crawled caption, is then input into the text encoder, and a multiple-positive contrastive loss will be applied. This can be expressed as:
\begin{equation}
L_{\text{contrast}} = - \frac{1}{2N} \sum_{i=1}^{N} \left( L_{\text{orig}}^i + L_{\text{syn-short}}^i \right)
\end{equation}
where
\begin{align}
L_{\text{orig}}^i &= \log \frac{\exp(S_{i, \text{orig}}^i)}{\sum_{j=1}^{N} \exp(S_{i, \text{orig}}^k)}, \\
L_{\text{syn-short}}^i &= \log \frac{\exp(S_{i, \text{syn-short}}^i)}{\sum_{j=1}^{N} \exp(S_{i, \text{syn-short}}^k)}
\end{align}
where \( S_{i, \text{orig}}^i \) and \( S_{i, \text{syn-short}}^i \) represent the scaled similarity for the original and short synthetic captions, respectively.

Note this training process is closely related to the prior work DreamLip~\cite{zheng2025dreamlip}---which uses the sub-caption strategy to preprocess synthetic data---but in a much simpler format. Specifically, in DreamLip, an average contrastive loss is calculated over all captions (\ie, the original web-crawled caption + a set of diverse synthetic captions) for image-text alignment. In comparison, ours directly trains with a single short synthetic caption rather than a set of them (\eg, typically this set contains more than 6 elements). As empirically shown in Section \ref{sec:exp}, our simplified strategy is sufficient to achieve strong performance, with our best model setting new state-of-the-art results in cross-modal retrieval tasks. In addition to this training simplification, our experiments are conducted at a significantly larger computational scale, \ie, our training dataset is $\app33\times$ larger (\ie,  Merged-30M~\cite{zheng2025dreamlip} \vs DataComp-1B) and the explored model size is up to ViT-H (while only up to ViT-B in DreamLip).

\begin{table*}[h!]
\centering
\caption{\textbf{Zero-shot cross-modal retrieval results on MSCOCO and Flickr30K}, with CLIPA and CoCa results reproduced by us. Both methods are implemented with a mixture training, where the original caption accounts for 80\% and the synthetic caption accounts for 20\%.}
\vspace{-.5em}
\resizebox{\textwidth}{!}{
\begin{tabular}{llcccccccccccc}
\toprule
\multirow{3}{*}{\textbf{Model}} & \multirow{3}{*}{\textbf{Method}} & \multicolumn{6}{c}{\textbf{MSCOCO}} & \multicolumn{6}{c}{\textbf{Flickr30K}} \\
\cmidrule(lr){3-8} \cmidrule(lr){9-14}
& & \multicolumn{3}{c}{Image$\rightarrow$Text} & \multicolumn{3}{c}{Text$\rightarrow$Image} & \multicolumn{3}{c}{Image$\rightarrow$Text} & \multicolumn{3}{c}{Text$\rightarrow$Image} \\
& & R@1 & R@5 & R@10 & R@1 & R@5 & R@10 & R@1 & R@5 & R@10 & R@1 & R@5 & R@10 \\
\midrule
\multirow{3}{*}{S/16} & CLIPA~\cite{li2024if}       & 55.8 & 79.5 & 87.6 & 35.0 & 60.9 & 71.6 & 79.6 & 95.4 & 98.0 & 59.2 & 83.2 & 89.3 \\
& CoCa~\cite{yu2022coca}   & 55.5 & 79.2 & 86.8 & 35.7 & 61.3 & 71.6 & 79.4 & 94.1 & 97.3 & 59.3 & 83.4 & 89.7 \\
                      &  \cellcolor{mygray} CLIP\texttt{S}   & \cellcolor{mygray}\textbf{61.8} & \cellcolor{mygray}\textbf{83.7} & \cellcolor{mygray}\textbf{90.0} & \cellcolor{mygray}\textbf{39.4} & \cellcolor{mygray}\textbf{65.2} & \cellcolor{mygray}\textbf{75.2} & \cellcolor{mygray}\textbf{85.9} & \cellcolor{mygray}\textbf{96.7} & \cellcolor{mygray}\textbf{98.5} & \cellcolor{mygray}\textbf{66.4} & \cellcolor{mygray}\textbf{87.2} & \cellcolor{mygray}\textbf{92.7} \\
                      \midrule
\multirow{3}{*}{B/16} & CLIPA~\cite{li2024if}       & 62.8 & 85.0 & 91.0 & 42.7 & 67.6 & 77.5 & 86.7 & 97.9 & 98.7 & 67.9 & 88.8 & 92.8 \\
& CoCa~\cite{yu2022coca}   & 63.1 & 84.4 & 90.1 & 43.2 & 68.3 & 77.6 & 87.8 & 97.8 & 99.2 & 68.2 & 88.9 & 93.2 \\
                      &  \cellcolor{mygray}CLIP\texttt{S}    & \cellcolor{mygray}\textbf{68.8} & \cellcolor{mygray}\textbf{88.4} & \cellcolor{mygray}\textbf{92.9} & \cellcolor{mygray}\textbf{48.2} & \cellcolor{mygray}\textbf{72.8} & \cellcolor{mygray}\textbf{81.4} & \cellcolor{mygray}\textbf{92.5} & \cellcolor{mygray}\textbf{99.3} & \cellcolor{mygray}\textbf{99.8} & \cellcolor{mygray}\textbf{75.3} & \cellcolor{mygray}\textbf{92.4} & \cellcolor{mygray}\textbf{95.8} \\
                      \midrule
\multirow{3}{*}{L/16} & CLIPA~\cite{li2024if}       & 66.8 & 87.0 & 92.5 & 47.8 & 72.6 & 81.3 & 91.3 & 98.9 & 99.4 & 73.9 & 92.2 & 95.6 \\
& CoCa~\cite{yu2022coca}   & 67.0 & 87.1 & 92.4 & 48.0 & 72.3 & 80.9 & 89.7 & 98.7 & 99.6 & 74.1 & 92.0 & 95.5 \\
                      & \cellcolor{mygray} CLIP\texttt{S}    & \cellcolor{mygray}\textbf{73.6} & \cellcolor{mygray}\textbf{90.6} & \cellcolor{mygray}\textbf{94.5} & \cellcolor{mygray}\textbf{53.6} & \cellcolor{mygray}\textbf{76.6} & \cellcolor{mygray}\textbf{84.4} & \cellcolor{mygray}\textbf{94.2} & \cellcolor{mygray}\textbf{99.2} & \cellcolor{mygray}\textbf{99.9} & \cellcolor{mygray}\textbf{80.6} & \cellcolor{mygray}\textbf{95.0} & \cellcolor{mygray}\textbf{97.2} \\
\bottomrule
\end{tabular}
}
\label{tab:retrieval_results}
\end{table*}

\subsection{Decoder: Predicting Full Synthetic Caption}
In our encoder design, we utilize only a portion of the synthetic captions during contrastive learning to achieve stronger performance and higher computation efficiency. However, this approach leaves substantial contextual information within the richly described synthetic captions unexploited. Furthermore, the preference for learning with shorter captions may suggest that the text encoder is unable to fully comprehend long sequences within the CLIP's contrastive learning framework. To this end, our second design introduces a generative task as an auxiliary objective to assist CLIP in capturing the full data distribution. Specifically, we adopt the caption modeling strategy from CoCa~\cite{yu2022coca} by incorporating an autoregressive decoder for caption generation. Importantly, unlike the symmetric decoder design in CoCa---where the decoder processes identical input and output text---we introduce an asymmetric learning structure: the text decoder takes the web-crawled caption as input and generates the full-length synthetic caption as the target. This learning approach mimics the recaptioning process performed by MLLMs, ensuring the full utilization of the knowledge contained in the synthetic captions. Additionally, it facilitates modeling the relationship between the two types of captions, with synthetic captions serving as a more semantically aligned and knowledge-rich representation of the web-crawled captions.

In our implementation, unlike the text branch in CoCa, we remove the causal mask from the original unimodal text decoder to implement a more effective text encoder, which will be further discussed in Sec.~\ref{sec:mask}. Then, in the multimodal text decoder, since the bidirectional text encoder's output cannot be directly used as input, we replace the sequential input text with a set of randomly initialized learnable tokens and use images and web-crawled captions as conditioning information. Formally, given the image features \( \mathbf{I}_{\text{image}} \) and learned tokens \( \mathbf{L}_{\text{learnable}} \), predicting full-length synthetic captions from web-crawled captions can be expressed based on the Eq.~\eqref{equation:generative loss} as:
\begin{equation}
L_{\text{gen}} = - \sum_{t=1}^{T} \log P(y_t \mid \mathbf{I}_{\text{image}}, \mathbf{C}_{\text{web}}, \mathbf{L}_{\text{learnable}, <t})
\end{equation}
where \( y_t \) represents the target token at time step \( t \), \( \mathbf{L}_{\text{learnable}, <t} \) denotes all learnable tokens preceding \( t \), and \( \mathbf{C}_{\text{web}} \) represents the web-crawled captions. The term \( P(y_t \mid \mathbf{I}_{\text{image}}, \mathbf{C}_{\text{web}}, \mathbf{L}_{\text{learnable}, <t}) \) is the conditional probability of generating the target token \( y_t \) given the image features, web-crawled captions, and preceding learnable tokens.

Moreover, our experiments show that simply concatenating information from different modalities yields stronger results than employing modality fusion with cross-attention mechanisms. Therefore, our decoder utilizes a self-attention mechanism accompanied by a specially designed combination mask. Specifically, to construct the input to the decoder, we concatenate the image features and the web-crawled caption to form the conditioning sequence, then concatenate this condition with the learnable tokens to form the complete input sequence. 
We design the combination mask \( M \) to ensure that tokens within the condition can attend to each other, while the learnable tokens follow an autoregressive prediction pattern. The mask \( M \) is defined as:
\begin{equation}
    M[i,j] = \begin{cases}
    1 & \text{if } i, j \leq L_{\text{cond}} \\
    1 & \text{if } i > L_{\text{cond}} \text{ and } j \leq L_{\text{cond}} \\
    1 & \text{if } i, j > L_{\text{cond}} \text{ and } i \geq j \\
    0 & \text{otherwise}
\end{cases}
\end{equation}
where \( L_{\text{cond}} \) represents the total length of the condition tokens, which are formed by concatenating the image tokens and the web-crawled caption tokens.

The combined input sequence passes through a self-attention mechanism guided by a mask \( M \) to enable appropriate token interactions. To align the generated sequence with the full-length synthetic captions \( \mathbf{T}_{\text{synthetic, full}} \), we compute the loss based on the likelihood of each token in the target generated sequence. Accordingly, the final generative loss \( L_{\text{gen}} \) is defined as:
\begin{equation}
    L_{\text{gen}} = - \sum_{t=1}^{T} \log P_\theta\left(y_t \mid \mathbf{I}_{\text{image}} \oplus \mathbf{C}_{\text{web}} \oplus \mathbf{L}_{\text{learnable}, <t}, M\right)
\end{equation}
where \( y_t \) represents the \( t \)-th token in the full-length synthetic caption \( \mathbf{T}_{\text{synthetic, full}} \).
The overall optimization objective of the entire model can be formalized as:
\begin{equation}
    L_{\text{total}} = \alpha \cdot L_{\text{contrast}} + \beta \cdot L_{\text{gen}}
\end{equation}
where \( \alpha \) and \( \beta \) are weighting factors that balance the contributions of the contrastive loss and the generative loss.

\section{Experiments}
\label{sec:exp}
\subsection{Pre-training Details}
We conduct our experiments using the Recap-DataComp-1B dataset~\cite{li2024if} for pre-training. Following the efficient training recipe introduced in \cite{li2024inverse}, all main experiments involve pre-training models for 2,000 ImageNet-equivalent epochs (\app2.6B samples seen) with images at a low resolution (\eg, resizing to $112\times112$ by default), followed by fine-tuning for 100 ImageNet-equivalent epochs at the resolution of $224\times224$. 
For the text branch, the input token length is 80,\footnote{Note that, despite setting the max text length as 80, we still only sample a single sentence from the synthetic caption as described in Section \ref{sec:encoder} and pad the remaining positions. Empirically, we observe that this strategy yields an additional \app1\% performance improvement compared to setting the maximum text length to 32.} and the output token length is 128, matching the number of learnable tokens in the decoder. 
The batch size is 32,768 for pre-training and 16,384 for fine-tuning. In the decoder’s autoregressive generation, image tokens and text tokens are concatenated to fuse information from different modalities. The weights for the contrastive loss (\(\alpha\)) and generative loss (\(\beta\)) are set to 1 and 2, respectively.

To further enhance training, in our experiment about comparison with state-of-the-art (SOTA) methods, we increase the pre-training epochs to 10,000 ImageNet-equivalent epochs (\app13B samples seen) and fine-tune for 400 ImageNet-equivalent epochs. The image resolution during pre-training is set to 84 for accelerating training, and remains at 224 during fine-tuning. Other settings remain consistent with those in the main experiments. More Detailed training parameters are provided in the appendix.

\begin{table*}[t!]
    \centering
    \caption{\textbf{Comparison with other SOTA vision-language pre-training methods} trained on public or private dataset. We report top-1 ImageNet-1K \cite{deng2009imagenet} classification accuracy and zero-shot recall of image and text retrieval on MSCOCO and Flickr30K.
}
\vspace{-.5em}
    \resizebox{\linewidth}{!}{
    \begin{tabular}{c|c|c|c|c|c|c|c|c|c}
    \toprule
       \multirow{2}{*}{\textbf{method}} & \multirow{2}{*}{\textbf{model size}} & \multirow{2}{*}{\textbf{\# patches}} & \multirow{2}{*}{\textbf{dataset}} & \multirow{2}{*}{\textbf{public}} &{\textbf{IN-1K}}& \multicolumn{2}{c}{\textbf{MSCOCO R$@$1}} & \multicolumn{2}{c}{\textbf{Flickr30K R$@$1}} \\
    \cmidrule(r){6-10}
     &&&&&val.&\multicolumn{1}{c}{I$\rightarrow$T} &\multicolumn{1}{c}{T$\rightarrow$I} &\multicolumn{1}{c}{I$\rightarrow$T} &\multicolumn{1}{c}{T$\rightarrow$I} \\
    \midrule
    CLIP~\citep{radford2021learning} & Large & 256 & WIT-400M~\citep{radford2021learning} & \XSolidBrush &75.5 & 56.3 & 36.5 & 85.2 & 65.0 \\
    CoCa~\citep{yu2022coca} & Large & 256 & LAION-2B~\citep{schuhmann2022laion} & \XSolidBrush &75.6 & 62.9 & 45.7 & 88.4 & 74.3 \\
    CLIP~\citep{gadre2024datacomp} & Large & 256 & DataComp-1B~\citep{gadre2024datacomp} & \CheckmarkBold &79.2 & 63.3 & 45.7 & 89.0 & 73.4 \\
    OpenCLIP~\citep{openclip} & Large & 256 & LAION-2B~\citep{schuhmann2022laion} & \CheckmarkBold  &75.5 & 63.4 & 46.5 & 89.5 & 75.5 \\
    SigLIP~\citep{zhai2023sigmoid} & Large & 256 & WebLI-5B~\citep{chen2022pali} & \XSolidBrush &\textbf{80.5} & 70.8 & 52.3 & 91.8 & 79.0 \\
    CLOC~\cite{chen2024contrastive} & Large & 576 & WiT~\cite{wu2023mofi}+DFN-5B~\cite{fang2023data} & \XSolidBrush &80.1 & 74.8 & 54.4 & - & - \\
    \rowcolor{mygray}CLIP\texttt{S} & Large & 256 & Recap-DataComp-1B~\cite{li2024if} & \CheckmarkBold &78.5 & \textbf{75.5} & \textbf{55.6} & \textbf{96.5} & \textbf{82.3} \\
    \midrule
    CLIP~\citep{fang2023data} & Huge & 729 & DFN-5B~\citep{fang2023data} & \XSolidBrush &\textbf{84.4} & 71.9 & 55.6 & 94.0 & 82.0 \\
    SigLIP~\citep{zhai2023sigmoid} & SO(400M) & 729 & WebLI-5B~\citep{chen2022pali} & \XSolidBrush &83.1 & 72.4 & 54.2 & 94.3 & 83.0 \\
    CLOC~\cite{chen2024contrastive} & Huge & 576 & WiT~\cite{wu2023mofi}+DFN-5B~\cite{fang2023data} & \XSolidBrush &81.3 & 75.7 & 55.1 & - & - \\
    \rowcolor{mygray}CLIP\texttt{S} & Huge & 256 & Recap-DataComp-1B~\cite{li2024if} & \CheckmarkBold & 79.9 & \textbf{76.4} & \textbf{57.2} & \textbf{96.6} & \textbf{83.9} \\
    \bottomrule
    \end{tabular}}
    \label{tab:sota_results}
\end{table*}

\begin{table*}[t!]
\centering
\caption{Comparison of LLaVA-1.5 performance trained with our visual encoder versus CLIP’s visual encoder across multiple MLLM benchmarks. The visual encoder size is L/14, the LLM used is LLaMA-3, and all results are reproduced by us.}
\vspace{-.5em}
\resizebox{\linewidth}{!}{
\begin{tabular}{lccccccccc}
\toprule
ViT          & MME-Cognition & MME-Perception & MMMU & MM-VET & GQA  & ChartQA & POPE & NoCaps & TextVQA\\
\midrule
OpenAI-CLIP        & 295.4           & \textbf{1433.5}            & 37.8 & 34.7   & 57.5 & 13.2           & 83.8  & 92.1& 56.8\\
CLIP\texttt{S}        &    \textbf{326.4}        & 1416.6            & \textbf{38.2} & \textbf{35.8}   & \textbf{60.3} & \textbf{14.2}           & \textbf{86.7}  & \textbf{102.5}&\textbf{57.6}\\
\bottomrule
\end{tabular}
}
\label{table:benchmark}
\end{table*}

\subsection{Evaluation}
\paragraph{Zero-Shot cross-modal retrieval.}
We evaluate the zero-shot cross-modal retrieval performance across different model sizes on the MSCOCO~\cite{lin2014microsoft} and Flickr30K~\cite{plummer2015flickr30k} datasets. We use CLIPA and CoCa as baselines and follow \cite{li2024if} to train with a mixture of web-crawled captions (80\%) and synthetic captions (20\%). 

As reported in Table~\ref{tab:retrieval_results}, our method consistently achieves superior performance across all benchmarks and model sizes, yielding significant improvements over the baselines. Notably, these substantial gains enable our smaller models to match or even surpass the performance of larger models from previous works. For instance, our S/16 model attains results comparable to the B/16 models of CLIPA and CoCa, and our B/16 model reaches the performance level of their L/16 models. These findings demonstrate the efficacy of our approach in enhancing cross-modal representation learning.

\paragraph{Comparision with SOTA methods.}
In Table~\ref{tab:sota_results}, we further compare our method with state-of-the-art vision-language pre-training approaches, reporting top-1 accuracy on ImageNet-1K and zero-shot recall rates for image and text retrieval tasks on MSCOCO and Flickr30K.  At the model size of L/14, our CLIP\texttt{S} achieves recall@1 scores of 75.5 and 96.5 for text retrieval, and 55.6 and 82.3 for image retrieval on MSCOCO and Flickr30K, respectively. These results are significantly higher than those of the prior art SigLIP \cite{zhai2023sigmoid}, and they also surpass the concurrent work CLOC \cite{chen2024contrastive}, which pretrains CLIP with a more fine-grained region-text contrastive loss. This performance trend remains consistent at the huge model size---\ie, with ViT-H/14, our CLIP\texttt{S} strongly attains recall@1 scores of 76.4 and 96.6 for text retrieval, and 57.2 and 83.9 for image retrieval, setting new SOTA records. 

Despite the strong cross-modal retrieval performance, we note that our CLIP\texttt{S} is less competitive on ImageNet zero-shot classification accuracy. We conjecture that this is mainly due to two factors: (1) training with synthetic caption is expected to yield lower ImageNet accuracy, as empirically shown in \cite{li2024if}; and (2) our training dataset, DataComp-1B, is expected to yield lower ImageNet accuracy compared to the higher-quality DFN dataset~\cite{fang2023data} and possibly Google's in-house WebLI dataset \cite{chen2022pali}.

\paragraph{CLIP\texttt{S} in LLaVA.}
To assess the visual representation capabilities of our pre-trained model, we integrate the CLIP\texttt{S} visual encoder into LLaVA-1.5~\cite{liu2024improved} and evaluate its performance. Specifically, we replace the original OpenAI-CLIP-L/14 visual encoder with our CLIP\texttt{S}-L/14 and utilized LLaMA-3~\cite{dubey2024llama} as the language model.
Since our pre-training employs images at a smaller resolution, we fine-tune CLIP\texttt{S}-L/14 at a resolution of $336 \times 336$ to match the configuration of OpenAI-CLIP-L/14. We then evaluate LLaVA's performance on multiple multimodal large language model (MLLM) benchmarks, including MME~\cite{fu2023mme}, MMMU~\cite{yue2023mmmu}, GQA~\cite{hudson2018gqa}, ChartQA~\cite{masry-etal-2022-chartqa}, POPE~\cite{Li-hallucination-2023}, NoCaps~\cite{agrawal2019nocaps}, and TextVQA~\cite{singh2019towards}.

The results summarized in Table~\ref{table:benchmark} demonstrate that integrating CLIP\texttt{S} significantly enhances LLaVA's performance across multiple metrics compared to using the original OpenAI-CLIP visual encoder. Specifically, out of the nine metrics evaluated, substituting OpenAI-CLIP with our CLIP\texttt{S} leads to performance gains in eight metrics. Notably, the most substantial improvements are observed on the NoCaps task, where performance increases by 10.4 points (from 92.1 to 102.5), and on the MME-Cognition task, with a boost of 31.0 points (from 295.4 to 326.4). Collectively, these results confirm that the strong visual capabilities of our CLIP\texttt{S} effectively transfer to the multimodal setting, enabling MLLMs to achieve a deeper understanding of visual content.

\subsection{Ablation Study}

\paragraph{Components.}
We hereby ablate the effects of different design components---sub-caption strategy, multi-positive contrastive loss, and generative loss---in CLIP\texttt{S}. We use the Recap-CLIP-B/16 model with a mixed ratio of 0.6~\cite{li2024if} as our baseline and examine how these components contribute to improvements in cross-modal retrieval performance on the MSCOCO dataset and classification accuracy on ImageNet.
We first introduce sub-caption extraction for the remaining 40\% of synthetic captions under the given setting. As reported in the second row of Table~\ref{tab:components}, this strategy enhances the model's cross-modal retrieval performance; but it also leads to a slight accuracy drop on ImageNet, possibly due to the reduced informational content in the synthetic sub-captions. Subsequently, we incorporate the multi-positive contrastive loss. By engaging in contrastive learning with both the original captions and the synthetic sub-captions, this approach facilitates the establishment of a more robust one-to-many relationship between images and captions. As a result, we observe substantial and consistent improvements in both cross-modal retrieval performance and ImageNet accuracy. Finally, we introduce the generative loss, which involves reconstructing the full synthetic captions based on the images and the original web-crawled captions. This addition consistently benefits all evaluated metrics. Specifically, when compared to the original baseline, this final configuration yields an improvement of +5.7\% in image-to-text retrieval, +5.1\% in text-to-image retrieval, and +1.2\% in ImageNet accuracy.

\begin{table}[t!]
\centering
\caption{Ablation study on components (Zero-shot Performance): SC = Sub-caption, MP = Multi-positive Contrastive Loss, GL = Generative Loss. I$\rightarrow$T = Image-to-Text Retrieval, T$\rightarrow$I = Text-to-Image Retrieval, both are MSCOCO's R@1.}
\vspace{-.5em}
\resizebox{.92\linewidth}{!}{
\begin{tabular}{l|c|c|c}
\toprule
\textbf{Component} & \textbf{I$\rightarrow$T} & \textbf{T$\rightarrow$I} & \textbf{IN1K} \\
\midrule
Baseline & 63.1 & 43.1 & 69.0 \\
+ SC & 64.5\red{$_{+1.4\%}$} & 44.7\red{$_{+1.6\%}$} & 68.4\textcolor{gray}{$_{-0.6\%}$} \\
+ SC \& MP & 67.1\red{$_{+4.0\%}$} & 46.0\red{$_{+2.9\%}$} & 69.4\red{$_{+0.4\%}$} \\
+ SC \& MP \& GL & 68.8\red{$_{+5.7\%}$} & 48.2\red{$_{+5.1\%}$} & 70.2\red{$_{+1.2\%}$} \\
\bottomrule
\end{tabular}
}
\label{tab:components}
\end{table}

\begin{figure}[t!]
    \centering
    \begin{subfigure}[b]{0.23\textwidth}
        \centering
        \includegraphics[width=\textwidth]{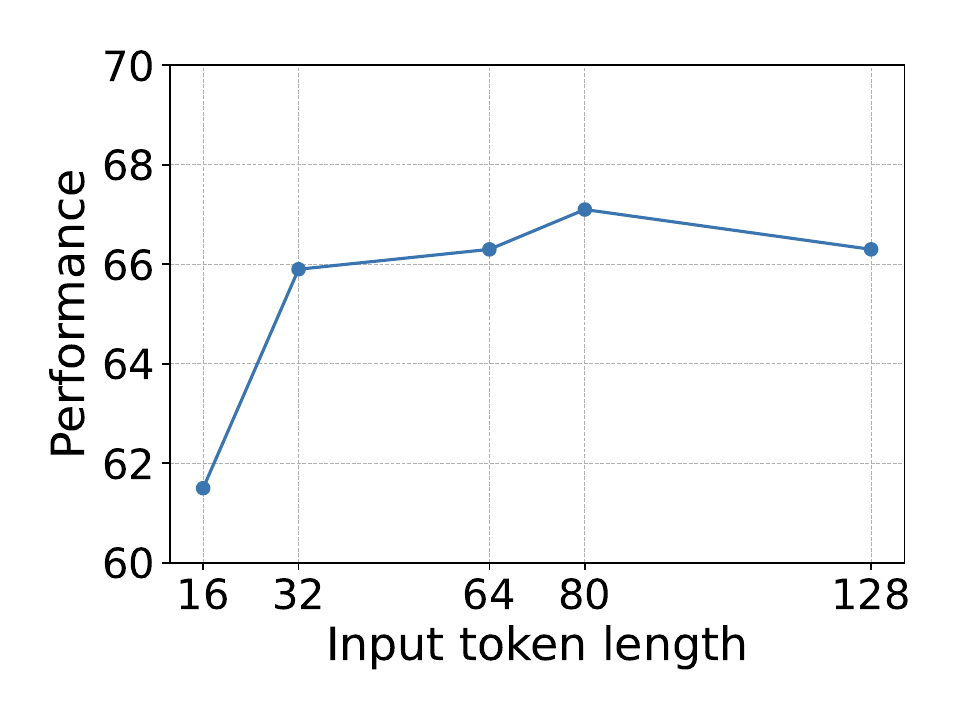}
        \caption{Input Token Length vs. Model Performance.}
        \label{fig:input_length}
    \end{subfigure}
    \hfill
    \begin{subfigure}[b]{0.23\textwidth}
        \centering
        \includegraphics[width=\textwidth]{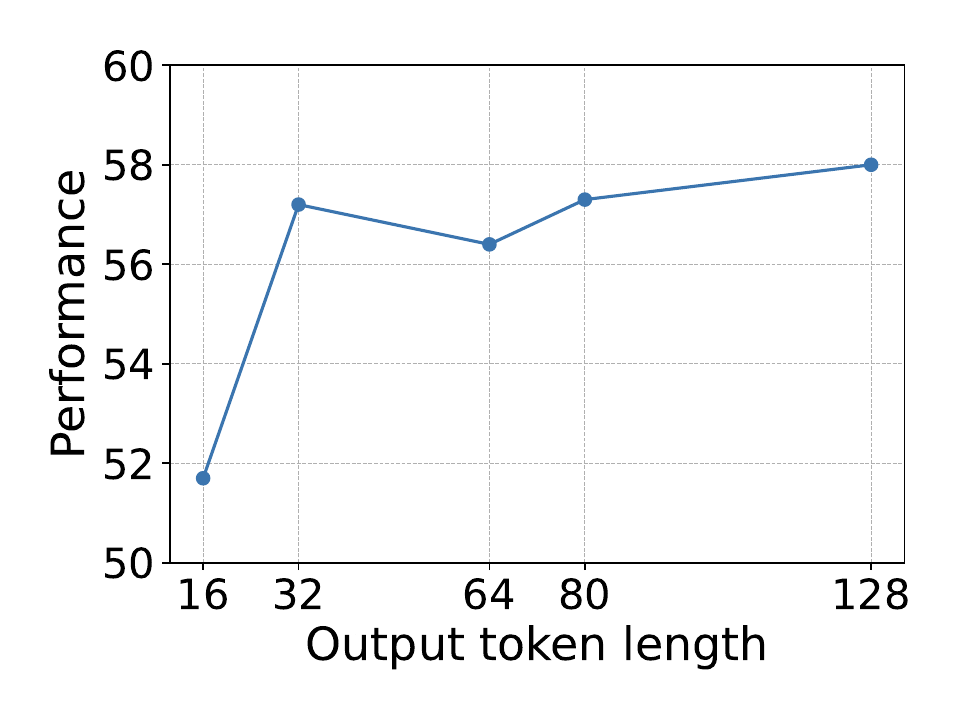}
        \caption{Output Token Length vs. Model Performance.}
        \label{fig:output_length}
    \end{subfigure}
    \vspace{-.5em}
    \caption{Ablation study on input and output token lengths. (a) pads a single sub-caption to different input lengths. (b) keeps the input valid token length constant and varies the target output token length. Performance is measured by R@1 of I→T on MSCOCO.}
    \label{fig:token_length}
\end{figure}

\paragraph{Single sub-caption.}
In Section~\ref{fig:scaling law}, we observe that the sequence length of the input text is a key factor affecting contrastive learning performance, with semantic continuity also playing a role. As randomly sampling multiple sub-captions and then truncating them can introduce semantic discontinuity, we hereby explore how randomly sampling a single sub-caption and padding it to different lengths affects model performance. As shown in Figure~\ref{fig:input_length}, increasing the input sequence length leads to improvements in model performance up to a certain point. This suggests that contrastive learning benefits from longer effective input lengths and that moderate padding can enhance performance. However, when the number of input tokens reaches 128, performance begins to decline, indicating that excessive padding may interfere with the model's ability to learn effective representations.

\paragraph{Generation target.}
\begin{table}[t!]
\centering
\caption{Ablation study on generation target (Zero-shot R@1 Retrieval Performance). Synthetic captions bring more significant performance improvements when used as generation targets.}
\vspace{-.5em}
\label{table:gen_target}
\resizebox{\linewidth}{!}{
\begin{tabular}{ccccc}
\toprule
\multirow{2}{*}{\textbf{Generation Target}} & \multicolumn{2}{c}{\textbf{MSCOCO}} & \multicolumn{2}{c}{\textbf{Flickr30K}} \\
\cmidrule(lr){2-3} \cmidrule(lr){4-5}
                                   & I$\rightarrow$T & T$\rightarrow$I & I$\rightarrow$T & T$\rightarrow$I \\
\midrule
Web captions                        & 55.9             & 36.9             & 88.1             & 85.0 \\
Synthetic captions                  & 58.4\red{$_{+2.5\%}$}             & 38.8\red{$_{+1.9\%}$}             & 90.2\red{$_{+2.1\%}$}             & 87.3\red{$_{+2.3\%}$} \\
\bottomrule
\end{tabular}
}
\end{table}
We explore how using text from different sources as generation targets affects model performance, as presented in Table~\ref{table:gen_target}. In these experiments, we maintain the use of web-crawled captions for contrastive learning but introduce an additional text batch for varying the generation target. Compared to the default setup in CoCa, where the web-crawled caption is set as the prediction target, our findings reveal that switching to synthetic captions as generation targets leads to significant performance improvements in cross-modal retrieval. This result underscores the rationale for selecting complete synthetic captions as prediction targets: their richer knowledge and smoother semantic distribution make them ideal for modeling, thereby achieving stronger performance gains.

\paragraph{Generated sequence length.}
We further examine how varying the lengths of generated sequences impact model performance while keeping the effective input length fixed. Due to the short length of the original captions, we rely entirely on synthetic captions to train CoCa.To maintain consistency in contrastive learning, we sample a single sub-caption and pad it to different lengths to preserve the effective text length, while exploring the impact of output lengths by adjusting the autoregressive label length. The experimental results are presented in Figure~\ref{fig:output_length}. We find that model performance tends to improve as the generated sequence length increases. This observation further confirms that long sequences are suitable for generative learning.

\paragraph{Fusion type and condition.}
In the context of generative learning, we hereby explore the impact of various conditioning modalities and fusion methods. Specifically, we employ concatenation and cross-attention as fusion types and examine the model's retrieval performance when conditioned on either the image alone or a combination of image and text. The experimental results are reported in Table~\ref{tab:generative_learning}.  Our findings indicate that integrating image and text---whether by using self-attention after concatenation or by applying cross-attention directly---consistently enhances model performance. Regarding the fusion techniques, cross-attention performs better when using the image alone as the condition, whereas concatenation slightly outperforms cross-attention when both image and text are used as conditions. Based on these results, we therefore choose concatenation as the default fusion type in our main experiments, with both image and text as conditions, to achieve the best performance.
\begin{table}[t!]
\centering
\caption{Ablation study on fusion types and conditions. Concat denotes concatenating condition tokens and learnable tokens as the input. Img\&Txt denotes concatenating image tokens and global text tokens as conditions.}
\vspace{-.5em}
\resizebox{0.9\linewidth}{!}{
\begin{tabular}{cc|cc|cc}
\toprule
\multicolumn{2}{c|}{\textbf{Fusion type}} & \multicolumn{2}{c|}{\textbf{Condition}} & \multicolumn{2}{c}{\textbf{MSCOCO}} \\
\cmidrule(r){1-2} \cmidrule(r){3-4} \cmidrule(r){5-6}
\textbf{Concat} & \textbf{Cross\_attn} & \textbf{Img} & \textbf{Img\&Txt} & \textbf{I→T} & \textbf{T→I} \\
\midrule
\checkmark &          & \checkmark &            & 67.8 & 47.1 \\
\checkmark &          &           & \checkmark & \textbf{68.8} & \textbf{48.2} \\
           & \checkmark & \checkmark &            & 68.3 & 47.4 \\
           & \checkmark &           & \checkmark & 68.3 & 48.2 \\
\bottomrule
\end{tabular}
}
\label{tab:generative_learning}
\end{table}
\paragraph{Causal mask.}
In CoCa, the text encoder uses a causal mask to ensure that the model processes text sequentially. However, we find that the causal mask is not essential in our CLIP\texttt{S}; in fact, removing it can enhance model performance. Specifically, in Table~\ref{tab:mask}, we explore the impact of the causal mask, learnable tokens, and input content on the model's efficacy. When utilizing the causal mask, we examine two settings: \emph{first prediction} and \emph{random prediction}. In the first prediction setting, the model uses the first sub-caption to predict the complete synthetic caption, whereas in the random prediction setting, it uses a randomly selected sub-caption. Our results indicate that, in both settings, the use of causal masks leads to a decrease in model performance. We hypothesize that this is because, in contrastive learning, capturing a strong global representation is more crucial than focusing on local features. This global representation also facilitates effective text reconstruction by the text decoder.

\begin{table}[t!]
\centering
\caption{Discussion about causal mask. L-tokens represents learnable tokens, and content represents the input text.}
\vspace{-.5em}
\resizebox{0.48\textwidth}{!}{
\begin{tabular}{cc|cc|cc|cc}
\toprule
\multicolumn{2}{c|}{\textbf{Causal mask}} & \multicolumn{2}{c|}{\textbf{L-tokens}} & \multicolumn{2}{c|}{\textbf{Content}} & \multicolumn{2}{c}{\textbf{MSCOCO}} \\
\cmidrule(r){1-2} \cmidrule(r){3-4} \cmidrule(r){5-6} \cmidrule(r){7-8}
\textbf{w/} & \textbf{w/o} & \textbf{w/} & \textbf{w/o} & \textbf{first} & \textbf{random} & \textbf{I→T} & \textbf{T→I} \\
\midrule
 &     \checkmark     & \checkmark &      & None &    None    & \textbf{68.8} & \textbf{48.2} \\
\checkmark &          &     \checkmark      &  & None& None &  67.8 & 47.6 \\
 \checkmark          &  &  &     \checkmark       & \checkmark&  &  66.8 & 46.2 \\
   \checkmark        &  &           & \checkmark & & \checkmark &  67.6 & 47.4 \\
\bottomrule
\end{tabular}
}
\label{tab:mask}
\end{table}
\label{sec:mask}

\section{Conclusion}
\label{sec:conclusion}
This work introduces CLIP\texttt{S}, with two simple and effective changes to enhance vision-language pre-training with synthetic captions. Our first design in on the text encoder---by observing the strong inverse effect in learning with short synthetic caption, we therefore feed only a portion of synthetic caption for contrastive learning. Then, to fully leverage the full synthetic caption, our second design incorporates an autoregressive captioner to mimic the recaptioning process---conditioning on the paired image input and web-crawled text description, the captioner learns to predict the full-length synthetic caption generated by advanced MLLMs. Experimental results show that CLIP\texttt{S} significantly improves zero-shot performance in cross-modal retrieval, reaching 76.4\% and 96.6\% for text retrieval and 57.2\% and 83.9\% for image retrieval on MSCOCO and Flickr30K, respectively, setting new SOTA results. Moreover, the visual encoder we trained strongly improves the visual capabilities of LLaVA, achieving notable gains across multiple MLLM benchmarks.

\appendix
\section{Experiment Details}
\label{sec:exp_details}
We provide more detailed pre-training hyperparameters and fine-tuning hyperparameters of main experiments in Table~\ref{tab:hyperparameters}. Except for the parameters related to our method, other parameters follow the settings in Recap-DataComp-1B~\cite{li2024if}. We provide the hyper-parameter settings of Large-14 and Huge-14 in the SOTA experiments in Table~\ref{tab:large14-hyperparameters}.
\begin{table}[ht]
\centering
\caption{Hyperparameters for Pretraining and Finetuning}
\vspace{-.5em}
\label{tab:hyperparameters}
\begin{tabular}{@{}lcc@{}}
\toprule
\textbf{Hyperparameter}          & \textbf{Pretraining} & \textbf{Finetuning} \\ \midrule
Learning rate                    & $8 \times 128 \times 10^{-6}$   & $4 \times 64 \times 10^{-7}$  \\
Batch size                       & 32768                  & 16384                 \\
Optimizer                        & \multicolumn{2}{c}{AdamW $(\beta_1=0.9, \beta_2=0.95)$}               \\
Weight decay                     & 0.2                 & 0.2              \\
Number of epochs                 & 2000                  & 100                  \\
Warm-up steps                    & 40                 & 20                 \\
CLIP-loss-weight                     & 1                  & 1                 \\
Caption-loss-weight           & 2                  & 2                 \\
Input token length               & 80                  & 80                 \\
Output token length              & 128                  & 128                 \\
Resolution               & 112                  & 224                 \\
temperature-init & 1/0.07                  & 1/0.07                 \\ \bottomrule
\end{tabular}
\vspace{-1em}
\end{table}

\begin{table}[ht]
\centering
\caption{Hyperparameters for SOTA Experiments}
\vspace{-.5em}
\label{tab:large14-hyperparameters}
\begin{tabular}{@{}lcc@{}}
\toprule
\textbf{Hyperparameter}          & \textbf{Pretraining} & \textbf{Fine-tuning} \\ \midrule
Learning rate                    & $8 \times 128 \times 10^{-6}$   & $4 \times 64 \times 10^{-7}$  \\
Batch size                       & 32768                          & 16384                          \\
Optimizer                        & \multicolumn{2}{c}{AdamW $(\beta_1=0.9, \beta_2=0.95)$}               \\
Weight decay                     & 0.2                            & 0.2                            \\
Number of epochs                 & 10000                           & 400                            \\
Warm-up steps                    & 40                             & 20                             \\
CLIP-loss-weight                 & 1                              & 1                              \\
Caption-loss-weight              & 2                              & 2                              \\
Input token length               & 80                             & 80                             \\
Output token length              & 128                            & 128                            \\
Resolution                       & 84                            & 224                            \\
temperature-init                 & 1/0.07                         & 1/0.07                         \\ \bottomrule
\end{tabular}
\end{table}

\section*{Acknowledgement}
We would like to thank TPU Research Cloud (TRC) program, Google Cloud Research Credits program, and AWS Cloud Credit for Research program for supporting our computing needs.

{
    \small
    \bibliographystyle{ieeenat_fullname}
    \bibliography{main}
}

\end{document}